# A Unified Framework of Bundle Adjustment and Feature Matching for High-Resolution Satellite Images

Xiao Ling, Xu Huang, and Rongjun Qin

**Abstract**

*Bundle adjustment (BA) is a technique for refining sensor orientations of satellite images, while adjustment accuracy is correlated with feature matching results. Feature matching often contains high uncertainties in weak/repeat textures, while BA results are helpful in reducing these uncertainties. To compute more accurate orientations, this article incorporates BA and feature matching in a unified framework and formulates the union as the optimization of a global energy function so that the solutions of the BA and feature matching are constrained with each other. To avoid a degeneracy in the optimization, we propose a comprised solution by breaking the optimization of the global energy function into two-step suboptimizations and compute the local minimums of each suboptimization in an incremental manner. Experiments on multi-view high-resolution satellite images show that our proposed method outperforms state-of-the-art orientation techniques with or without accurate least-squares matching.*

## Introduction

Georeferencing of multi-view satellite images is a necessary step to geometrically align these images in a common coordinate system (Qin 2016) and to allow subsequent applications in, for example, digital surface model (DSM) generation (Di Rita *et al.* 2017; Qin 2016, 2017, 2019), change detection (Zhuang *et al.* 2018), data fusion (Mohammadi *et al.* 2019), and so on. Typically, the georeferencing of multi-view images follows two major steps: (1) feature point matching across different images and (2) bundle adjustment (BA), which aligns the positions and attitudes of the cameras so that the optical rays from corresponding pixels intersect at the same ground point in the object space. However, the accuracy of BA relies highly on the quality of the feature matching results (e.g., point localization accuracy and distributions). Many of the existing feature point extraction methods, such as the scale-invariant feature transform (SIFT) (Lowe 2004) and Speeded Up Robust Features (Bay *et al.* 2006) operators, utilize the scale space by locating the interest point in a reduced resolution, and the descriptors of these operators explore the gradient of local patches and do not accommodate geometric distortions of the local patches (affine or perspective), which often lead to reduced location and matching accuracy of the points. Since the standard BA takes the matched feature points as observations and optimizes only the poses, we hypothesize that simultaneously optimizing the pose and the location of the matched points may improve georeferencing accuracy.

There have been several attempts (Dos Santos *et al.* 2016; Ling *et al.* 2016; Noh and Howat 2018; Zhou *et al.* 2018) to combine feature matching and BA for more accurate orientations. All of these works follow a coarse-to-fine strategy: they used either coarse BA results (from a few robust initial matches) or original sensor model parameters to reduce the searching space of each feature point with epipolar constraints, then found a large number of robust matches in the reduced searching space for the accurate BA. However, these methods did not fully utilize the BA results for the purpose of subpixel-level matching. Their epipolar constraints reduced the searching space only at the integer pixel level, thus resulting in only pixel-level matches. To achieve more accurate matching results, the least-squares matching (LSM) technique (Hu and Wu 2017) is used to find subpixel-level matches by estimating the best fit between matching windows of corresponding pixels. However, given the nonlinear and nonquadric nature of the LSM formation, the convergence of the LSM is often uncontrolled, especially in weak-textured or nonlinear radiometric distortion regions (Gruen 2012).

To further improve orientation accuracies, we combine feature matching and BA in a unified framework where the solutions of feature matching and BA are mutually constrained. The rationale here is (1) to improve point localization/matching accuracy and (2) to improve 2D and 3D consistency through the constraints casted by the use of an approximately adjusted rational function model and to reduce the chances of divergence in LSM. The core algorithm formulates the combination into the optimization of a global energy function where feature matching and BA are formulated as two terms in the energy function. However, the solutions of feature matching and BA have some redundant variables that may bring a degeneracy in the optimization. We therefore propose a comprised solution by breaking the optimization into suboptimizations and, respectively, assigning the redundant variables into different suboptimizations so that the variables in the suboptimizations are independent. Finally, the local minimums of the suboptimizations are computed in an incremental manner. Experiments on a multi-view satellite data set show that our approach is capable of computing more accurate orientation results with more accurate matches when compared with the state-of-the-art orientation techniques with or without LSM points.

Xiao Ling is with Ohio State University, Columbus, OH 43210.

Xu Huang is with Ohio State University, Columbus, OH 43210

Rongjun Qin is with Ohio State University, Columbus, OH 43210 (qin.324@osu.edu).

## Methodology

### Problem Formulation

Given multi-view high-resolution satellite images $i$, corresponding generic sensor models (rational polynomial coefficients [RPC] [Grodecki and Dial 2003]) $R$, and a series of multi-view matches $\mathbf{P} = \{P_i^j\}$ with $P_i^j$ being an image matching point of the object space point $i$ on the image $j$, our goal is to estimate accurate BA results with corrected matches. In general, we consider the geometric constraints of the forward and backward projections (also called reprojections) in the BA and the photo-consistency constraints in the LSM and formulate these two constraints as two terms in a global energy function, as shown in Equation 1. The first term, $e_{proj}$, formulates the reprojection error, which is the squared sum of distances between the corrected feature matching points and the reprojection points, and the second term, $e_{LSM}$, is the LSM objective term, which measures the similarities in intensity of corresponding matching windows against the linear radiometric distortions. These two terms are mutually constrained since the reprojection error $e_{proj}$ uses measurement from the LSM $(\Delta x_i^j, \Delta y_i^j)^T$ and the LSM error term $e_{LSM}$ takes the refined parameters from the reprojection error term $(x_0^j, y_0^j)^T$:

(Equation 1, *see below*)

where $E$ is the global energy function; $\mathbf{x}_0, \mathbf{y}_0$ are vectors of RPC biases of all satellite images in column and row directions; $\Delta \mathbf{x}, \Delta \mathbf{y}$ are vectors of image coordinate corrections of the given feature matching points; $\mathbf{N}, \mathbf{L}, \mathbf{H}$ are vectors of object space point coordinates of all matches in latitude, longitude, and height directions; $\mathbf{h}_0, \mathbf{h}_1$ are vectors of radiometric distortion corrections that are used to correct the linear intensity distortions among multi-view matches; $i$ is an object space point; $\mathbf{S}_i$ is a set of images that contain the matches of $i$; $j$ is an image in $\mathbf{S}_i$; $\mathbf{R}_j = (R_{j,x}, R_{j,y})^T$ is the RPC model of the image $j$ in column and row directions; $(N_i, L_i, H_i)^T$ is ground coordinates (latitude, longitude, and height) of the object space point $i$; $S_{s,j}, O_{s,j}, S_{l,j}, O_{l,j}$ are default RPC normalization parameters of the image $j$ in the column and row directions; $x_0^j, y_0^j$ are RPC biases of the image $j$ in the column and the row directions; $\bar{R}_j(N_i, L_i, H_i) - (x_0^j, y_0^j)^T$ is the biased reprojection points of the object space point $i$ on the image $j$; $\Delta x_i^j, \Delta y_i^j$ are image coordinate corrections of $P_i^j$; and $b$ is a reference image so that the matches in other images in $\mathbf{S}_i$ must satisfy the photo-consistency with $b$. Therefore, the image coordinate corrections $\Delta x_i^b, \Delta y_i^b$ can be set as 0. In this article, we compute zero-based normalized cross-correlation scores for any two corresponding pixels in the matches, accumulate the scores that are related to the same image, and find an image with the highest accumulation results as the reference image; $h_{i,0}^j, h_{i,1}^j$ are radiometric distortion corrections of $P_i^j$, and $\mathbf{I}_{j,w}$ is a matching window on the image $j$ with the window size $w$.

Often, on-orbit geometric calibration has compensated for high-order systematic errors; thus, orientation errors can be corrected by constant biases in the image space as long as the corresponding satellite orbit is no longer than 500 km (Fraser and Hanley 2005). Therefore, we adopt the constant RPC biases in Equation 1. We also adopt the LSM model (Hu and Wu 2017) with eight parameters (two radiometric parameters and six geometric parameters) to correct image coordinates of matches. Thus, the global energy function in Equation 1 can be minimized through standard Newton-Raphson method, as shown in Equation 2. The first two equations are derived from the geometric term, and the third equation is derived from the photo-consistency term:

$$f_y^{i,j} = \frac{y'_i^j + \Delta y_i^j + y_0^j - O_{l,j}}{S_{l,j}} - R_{j,y}(N_i, L_i, H_i)$$

$$f_x^{i,j} = \frac{x'_i^j + \Delta x_i^j + x_0^j - O_{s,j}}{S_{s,j}} - R_{j,x}(N_i, L_i, H_i) \quad j \in \mathbf{S}_i$$

$$f_{c,\delta}^{i,j} = \mathbf{I}_b(P_i^b + \delta) - h_{i,0}^j - h_{i,1}^j \mathbf{I}_j\left(P_i^j + \delta + (\Delta x_i^j, \Delta y_i^j)^T\right) \quad \begin{array}{l} j, b \in S_i \\ j \neq b \end{array} \quad (2)$$

$$\Delta y_i^j = b0_i^j + b1_i^j \cdot x'_i^b + b2_i^j \cdot y'_i^b$$

$$\Delta x_i^j = a0_i^j + a1_i^j \cdot x'_i^b + a2_i^j \cdot y'_i^b$$

$$\Delta x_i^b = \Delta y_i^b = 0$$

$$P_i^b + \delta \in \mathbf{I}_{b,w}(P_i^b) \quad P_i^j + \delta \in \mathbf{I}_{j,w}(P_i^j)$$

where $f_y^{i,j}, f_x^{i,j}$ are functions to measure the image location differences between the corrected image point of $P_i^j$ and its corresponding reprojection point; $x'_i^j, y'_i^j$ are column and row coordinates of $P_i^j$ in the image space; $P_i^j + \delta$ is a pixel in the matching window of $P_i^j$ with $\delta$ being an offset vector to $P_i^j$; $f_{c,\delta}^{i,j}$ is a function to measure the photo-consistency between $P_i^b + \delta$ and $P_i^j + \delta$; $x'_i^b, y'_i^b$ are column and row coordinates of $P_i^b$ in the reference image $b$; and $a0_i^j, a1_i^j, a2_i^j, b0_i^j, b1_i^j, b2_i^j$ are affine parameters to correct image coordinates of $P_i^j$. For each matching point $P_i^j$, the number of geometric equations $f_x^{i,j}$ and $f_y^{i,j}$ is only one, while the number of the photo-consistency equations $f_{c,\delta}^{i,j}$ are in total $w^2$. Therefore, it is better to assign higher weights to $f_x^{i,j}, f_y^{i,j}$ during the optimization so that the geometric term and the photo-consistency term can make comparable contributions to the final solution. The detailed weight strategy is introduced next.

### Solution

Equation 2 is nonlinear, so it is hard to directly compute the optimal solution. Therefore, we first estimate the initial values of all unknowns $\mathbf{x}_0, \mathbf{y}_0, \mathbf{N}, \mathbf{L}, \mathbf{H}, \Delta \mathbf{x}, \Delta \mathbf{y}, \mathbf{h}_0, \mathbf{h}_1$ and then linearize Equation 2 by the first-order Taylor expansion. The initial values of $\mathbf{x}_0, \mathbf{y}_0$ are set as 0; the initial values of $\mathbf{N}, \mathbf{L}, \mathbf{H}$ are computed by triangulating the given matches using the original RPC parameters; and the initial values of the image coordinate correction parameters $\{a0_i^j, a1_i^j, a2_i^j\}$ and $\{b0_i^j, b1_i^j, b2_i^j\}$ are, respectively, set as $\{-x'_i^b, 1, 0\}$ and $\{-y'_i^b, 1, 0\}$. We normalize the intensities in all matching windows so that the initial values of $\mathbf{h}_0, \mathbf{h}_1$ can be set as 0 and 1. Given these initial values, the linearization of Equation 2 is as follows:

$$V_y^{i,j} = \frac{\partial (f_y^{i,j})_0}{\partial y_0^j} dy_0^j + \frac{\partial (f_y^{i,j})_0}{\partial b0_i^j} db0_i^j + \frac{\partial (f_y^{i,j})_0}{\partial b1_i^j} db1_i^j + \frac{\partial (f_y^{i,j})_0}{\partial b2_i^j} db2_i^j + \frac{\partial (f_y^{i,j})_0}{\partial N_i} dN_i + \frac{\partial (f_y^{i,j})_0}{\partial L_i} dL_i + \frac{\partial (f_y^{i,j})_0}{\partial H_i} dH_i - \left(-(f_y^{i,j})_0\right)$$

$$V_x^{i,j} = \frac{\partial (f_x^{i,j})_0}{\partial x_0^j} dx_0^j + \frac{\partial (f_x^{i,j})_0}{\partial a0_i^j} da0_i^j + \frac{\partial (f_x^{i,j})_0}{\partial a1_i^j} da1_i^j + \frac{\partial (f_x^{i,j})_0}{\partial a2_i^j} da2_i^j + \frac{\partial (f_x^{i,j})_0}{\partial N_i} dN_i + \frac{\partial (f_x^{i,j})_0}{\partial L_i} dL_i + \frac{\partial (f_x^{i,j})_0}{\partial H_i} dH_i - \left(-(f_x^{i,j})_0\right)$$

$$V_{c,\delta}^{i,j} = \frac{\partial (f_{c,\delta}^{i,j})_0}{\partial h0_i^j} dh0_i^j + \frac{\partial (f_{c,\delta}^{i,j})_0}{\partial h1_i^j} dh1_i^j + \frac{\partial (f_{c,\delta}^{i,j})_0}{\partial a0_i^j} da0_i^j + \frac{\partial (f_{c,\delta}^{i,j})_0}{\partial a1_i^j} da1_i^j + \frac{\partial (f_{c,\delta}^{i,j})_0}{\partial a2_i^j} da2_i^j + \frac{\partial (f_{c,\delta}^{i,j})_0}{\partial b0_i^j} db0_i^j + \frac{\partial (f_{c,\delta}^{i,j})_0}{\partial b1_i^j} db1_i^j + \frac{\partial (f_{c,\delta}^{i,j})_0}{\partial b2_i^j} db2_i^j - \left(-(f_{c,\delta}^{i,j})_0\right) \quad (3)$$

$$j, b \in S_i$$
$$j \neq b$$
$$da0_i^b = da1_i^b = da2_i^b = db0_i^b = db1_i^b = db2_i^b = 0$$

---

$$\min E(x_0, y_0, N, L, H, \Delta x, \Delta y, h_0, h_1) = \sum_i e_{proj} + e_{LSM} = \sum_i \left( \sum_{j \in S_i} \left\| \bar{R}_j(N_i, L_i, H_i) - (x_0^j, y_0^j)^T - \left(P_i^j + (\Delta x_i^j, \Delta y_i^j)^T\right) \right\|_2 + \sum_{j, b \in S_i, j \neq b} \left\| \mathbf{I}_{b,w}(P_i^b) - h_{i,0}^j - h_{i,1}^j \mathbf{I}_{j,w}\left(P_i^j + (\Delta x_i^j, \Delta y_i^j)^T\right) \right\|_2 \right)$$

$$\bar{R}_j(N_i, L_i, H_i) = \left(S_{s,j} \cdot R_{j,x}(N_i, L_i, H_i) + O_{s,j}, S_{l,j} \cdot R_{j,y}(N_i, L_i, H_i) + O_{l,j}\right)^T \quad (1)$$

$$\Delta x_i^b = \Delta y_i^b = 0$$

where $(f_y^{i,j})_0$, $(f_x^{i,j})_0$, and $(f_{c,\delta}^{i,j})_0$ are initial values of functions $f_y^{i,j}$, $f_x^{i,j}$, and $f_{c,\delta}^{i,j}$; $V_x^{i,j}$, $V_y^{i,j}$ are geometric residual errors of the image locations of $P_i^j$; $V_{c,\delta}^{i,j}$ is the intensity residual error; and $dy_0^j$, $dx_0^j$, $da0_i^j$, $da1_i^j$, $da2_i^j$, $db0_i^j$, $db1_i^j$, $db2_i^j$, $dh0_i^j$, $dh1_i^j$, $dN_i$, $dL_i$, and $dH_i$ are corrections of all unknowns in the global energy function.

However, there exist some redundancies between the RPC biases $x_0^j$, $y_0^j$ and the image coordinate corrections $a0_i^j$, $b0_i^j$, which may make the residual error equations in Equation 3 degenerate. Moreover, the large quantities of the image coordinate corrections also bring great challenges in efficient computation. Therefore, we assign the RPC biases and the image coordinate corrections in different computation tasks and propose a comprised solution by breaking the optimization of the global energy function in Equation 1 into two kinds of suboptimizations in an incremental manner: (1) the suboptimization of RPC biases and object space point coordinates without considering image coordinate corrections or radiometric distortion corrections and (2) the suboptimization of image coordinate corrections, radiometric distortion corrections, and object space point coordinates with refined RPC biases from the first suboptimization.

In the first suboptimization, the image coordinate corrections and radiometric distortion corrections are fixed as the initial values, while only the RPC biases and the object space point coordinates are considered. Therefore, Equation 3 can be simplified by removing the terms of image coordinate corrections and radiometric distortion corrections in $V_y^{i,j}$, $V_x^{i,j}$ and removing the entire function of $V_{c,\delta}^{i,j}$ as follows:

$$V_y^{i,j} = \frac{\partial (f_y^{i,j})_0}{\partial y_0^j} dy_0^j + \frac{\partial (f_y^{i,j})_0}{\partial N_i} dN_i + \frac{\partial (f_y^{i,j})_0}{\partial L_i} dL_i + \frac{\partial (f_y^{i,j})_0}{\partial H_i} dH_i - \left(-(f_y^{i,j})_0\right)$$

$$V_x^{i,j} = \frac{\partial (f_x^{i,j})_0}{\partial x_0^j} dx_0^j + \frac{\partial (f_x^{i,j})_0}{\partial N_i} dN_i + \frac{\partial (f_x^{i,j})_0}{\partial L_i} dL_i + \frac{\partial (f_x^{i,j})_0}{\partial H_i} dH_i - \left(-(f_x^{i,j})_0\right)$$

(4)

Equation 4 is similar to the traditional BA model. Least-squares or gradient descent methods (Zheng and Zhang 2016) can be used to compute the optimal solution of Equation 4. The RPC biases are then used to refine the RPC parameters of each image, and the refined RPCs are used in the second suboptimizations to provide geometric guidance for more accurate matching. Given the refined RPC parameters, the second suboptimizations of Equation 3 can be simplified as follows:

(Equation 5, *see below*)

The initial values of the ground coordinates $N$, $L$, $H$ can be updated by the refined RPCs in the first suboptimization. Since RPC bias corrections were not considered in Equation 5, the second suboptimization can be formulated as the individual optimizations of each match. However, the solution of Equation 5 depends partly on the geometric accuracies of the refined RPCs. High-accuracy RPCs can provide accurate geometric guidance for the feature matching and vice versa. To compute robust solutions of Equation 5, we formulate the geometric accuracies of the refined RPCs as the reprojection errors of the matches and define the root-mean-square error (RMSE) of reprojection errors of the matches as weights for the geometric terms $V_y^{i,j}$ and $V_x^{i,j}$ in Equation 5. Higher reprojection errors correspond to lower weights and vice versa. To ensure that the geometric terms $V_y^{i,j}$, $V_x^{i,j}$ make comparable contributions with the photo-consistency term $V_{c,\delta}^{i,j}$ in the solution of Equation 5, we consider both the reprojection errors and the matching window sizes in the weight strategy as follows:

$$W_{reprj} = W_{max} \cdot \exp(-\varepsilon^2/\sigma)$$

$$W_{max} = P \cdot w^2 \cdot \frac{n-1}{2}$$

$$\varepsilon = \sqrt{\sum_j^n \left(d_{j,x}^2 + d_{j,y}^2\right)/(n-1.5)}$$

(6)

where $W_{reprj}$ is a weight of $V_y^{i,j}$, $V_x^{i,j}$, depending on the RMSE of reprojection errors and the matching window sizes; $\varepsilon$ is the RMSE of reprojection errors; $n$ is the viewing ray number of the matches; and $d_{j,x}$, $d_{j,y}$ are reprojection errors between the matching points and the reprojection points in the column and row directions on the image $j$. Since each matching point gives two observations and the unknowns of the object space point coordinates are only three, 1.5 was subtracted from the viewing ray number for more robust RMSE evaluation. $W_{max}$ means the maximum weights of $V_y^{i,j}$, $V_x^{i,j}$ in the solution. Since the number of the photo-consistency terms $V_{c,\delta}^{i,j}$ of each match is $w^2 \cdot (n-1)/2$ times larger than the number of the geometric terms, $W_{max}$ is scaled by $w^2 \cdot (n-1)/2$ to ensure comparable contributions of the geometric terms and the photo-consistency terms in the solution of Equation 5. $P$ is a factor that controls the maximum weights in the solution, which is empirically set as 0.5 in our experiments. $\sigma$ is a factor that controls the impact of reprojection errors on the weights of the geometric terms. $\sigma$ is set as two pixels in our experiments so that large reprojection errors (larger than two pixels) will produce negligible weights for the geometric terms. Therefore, the uncertainties in the refined RPCs have only negligible impacts on the matches correction process.

However, too small weights (close to zero) may make the error equations in Equation 5 degenerate; thus, we also add a virtual ground control point (VGCP) constraint by setting the refined object space point in the first suboptimizations as VGCP as follows:

$$dlat_i = 0 \quad dlon_i = 0 \quad dhei_i = 0 \quad (7)$$

$$W_{VGCP} = W_{max} - W_{reprj}$$

where $W_{VGCP}$ is a weight of VGCP constraints. If reprojection errors are small, the geometric terms $V_y^{i,j}$, $V_x^{i,j}$ made more contributions to the solution so that the feature matching corrections are geometrically constrained. Otherwise, the VGCP constraints made more contributions in the solution. In an extreme case that $W_{VGCP} \approx 1$, the solution of the second suboptimizations will degenerate into the traditional LSM model (Hu and Wu 2017).

Next, we compute the solution of the second suboptimizations of each match by combining Equations 5–7. The image

$$V_y^{i,j} = \frac{\partial (f_y^{i,j})_0}{\partial b0_i^j} db0_i^j + \frac{\partial (f_y^{i,j})_0}{\partial N_i} dN_i + \frac{\partial (f_y^{i,j})_0}{\partial L_i} dL_i + \frac{\partial (f_y^{i,j})_0}{\partial H_i} dH_i - \left(-(f_y^{i,j})_0\right)$$

$$V_x^{i,j} = \frac{\partial (f_x^{i,j})_0}{\partial a0_i^j} da0_i^j + \frac{\partial (f_x^{i,j})_0}{\partial N_i} dN_i + \frac{\partial (f_x^{i,j})_0}{\partial L_i} dL_i + \frac{\partial (f_x^{i,j})_0}{\partial H_i} dH_i - \left(-(f_x^{i,j})_0\right) \quad j \in \mathbf{S}_i$$

(5)

$$V_{c,\delta}^{i,j} = \frac{\partial (f_{c,\delta}^{i,j})_0}{\partial h0_i^j} dh0_i^j + \frac{\partial (f_{c,\delta}^{i,j})_0}{\partial h1_i^j} dh1_i^j + \frac{\partial (f_{c,\delta}^{i,j})_0}{\partial a0_i^j} da0_i^j + \frac{\partial (f_{c,\delta}^{i,j})_0}{\partial a1_i^j} da1_i^j + \frac{\partial (f_{c,\delta}^{i,j})_0}{\partial a2_i^j} da2_i^j + \frac{\partial (f_{c,\delta}^{i,j})_0}{\partial b0_i^j} db0_i^j + \frac{\partial (f_{c,\delta}^{i,j})_0}{\partial b1_i^j} db1_i^j + \frac{\partial (f_{c,\delta}^{i,j})_0}{\partial b2_i^j} db2_i^j - \left(-(f_{c,\delta}^{i,j})_0\right) \quad \begin{array}{l} j,b \in \mathbf{S}_i \\ j \neq b \end{array}$$

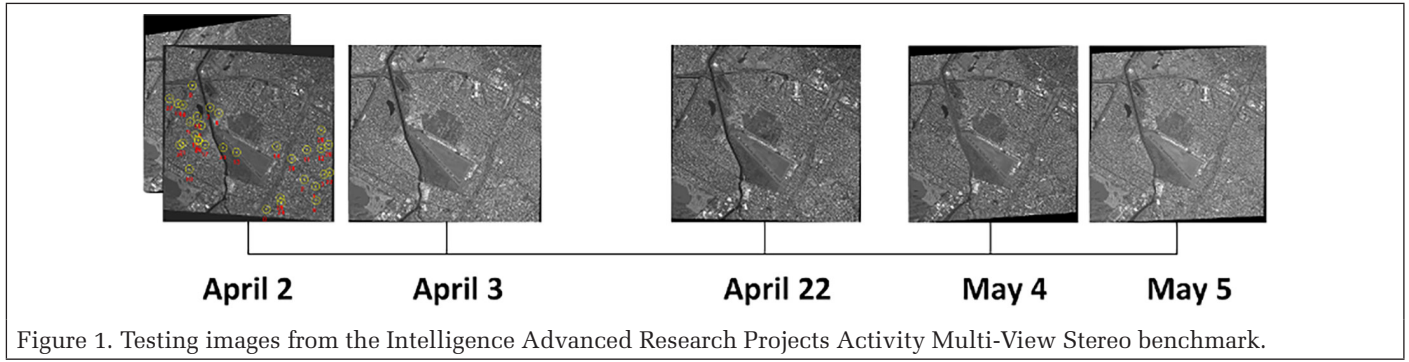

Figure 1. Testing images from the Intelligence Advanced Research Projects Activity Multi-View Stereo benchmark.

coordinate corrections in the solution are then used to correct the matches. Finally, the more accurate matching results and the corresponding refined object space points are used in the first suboptimizations for more accurate BA results.

## Experiments

Our proposed method was tested on six WorldView-3 panchromatic images near San Fernando, Argentina, from the Intelligence Advanced Research Projects Activity Multi-View Stereo benchmark data set (Bosch *et al.* 2016). The rationale of using the single-source data sets to test the performance of our methods is twofold. First, our methods can be used for any images that use rational function models and are not sensor specific, and the use of single-source data helps keep the experiments impacted by other factors, such as images with complex distortions. Second, these images can utilize observations with as many as six rays, which is sufficient to test a BA system while remaining manageable in an experimental setup. All images are Level 2 products, which means that high-order systematic errors in the ancillary data have already been compensated for and have been systematically mapped into a standard cartographic map projection based on a prediction of where the satellite was when the image was acquired (NASA 2019). Their sizes are around 13000×12000 and have a greater than 90% overlap ratio. The covering landscape is a typical urban area containing buildings, rivers, trees, and so on. The ground sampling distance of these images is about 0.5 m, and the imaging time of them was from April 2 to May 5, as shown in Figure 1.

To demonstrate the generality and reliability of the proposed method, we utilized a mainstream SIFT feature matching method to obtain initial matches. There are a total of 4210 SIFT matches (see Figure 2), more than 90% of which have the multi-view connectivity (at least three). Given these SIFT matches, our proposed method was compared with two methods: (1) a state-of-the-art orientation method (Ozcanli *et al.* 2014) that initially corrected RPC biases between stereo pairs, then removed mismatching image points/outliers by eliminating the points whose distances to their corresponding epipolar lines are larger than two pixels, and finally computed the BA results in a reference coordinate system of the best stereo pair without any matches corrections (termed BA), and (2) an LSM-based method that followed the same BA technique with the first method. However, the initial matches were corrected by the LSM technique (Hu and Wu 2017) before BA (termed LSM + BA). For fair comparisons, the outlier elimination technique is also applied in our proposed method after the image coordinate corrections in the second suboptimizations.

Both the proposed method and the LSM + BA need fixed matching windows to correct matches; we therefore utilized a series of different matching windows—{5×5, 7×7, 9×9, 11×11, 13×13, 15×15, 17×17, 19×19, 21×21, 31×31, 41×41}—in the proposed method and the LSM + BA for matches corrections, thus resulting in different BA results. We then compared the results of the three methods from four aspects: (1) the number of diverged matches in the LSM procedure (fewer diverged matches mean a more reliable LSM algorithm); (2) the number of outliers (fewer outliers mean a more robust matching result); (3) average RMSE of reprojection errors for the matches in BA and the corrected matches in LSM + BA and our proposed method (termed internal accuracy), which is used to evaluate the fitting between matches/corrected matches and the BA results; and (4) average RMSE of reprojection errors for the manually selected matches (termed external accuracy), which gives a more comprehensive evaluation to the final orientation accuracies.

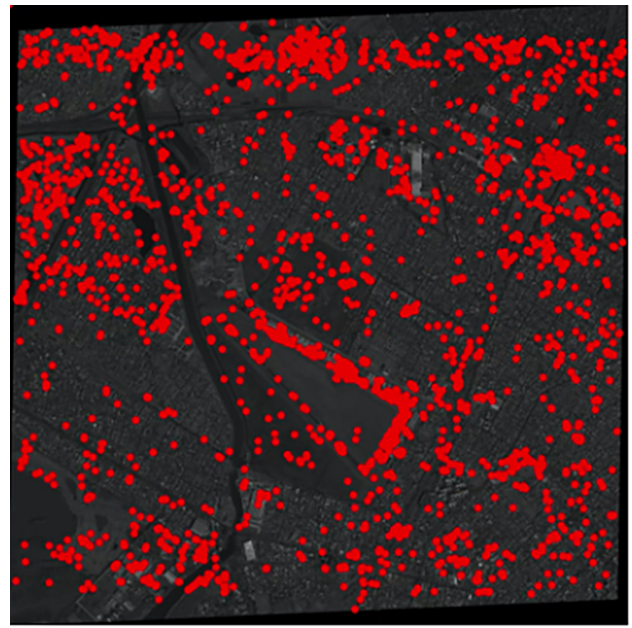

Figure 2. Distribution of scale-invariant feature transform (SIFT) matches on one image. All SIFT points are colored in red.

## Convergence of the LSM

The LSM is able to locate subpixel-level matches, while the convergence of the traditional LSM is often uncontrolled due to weak-textured or nonlinear radiometric distortion regions. To overcome this weakness, the proposed method not only introduces the photo-consistency term, as does the traditional LSM method, but also incorporates geometric terms into the LSM procedure to improve its convergence. A series of matching windows were utilized in the traditional LSM and the proposed method, then the numbers of the diverge matches were counted as an evaluation indicator to check the performance of the two methods. The output numbers of diverged matches are shown in Figure 3.

Figure 3 shows that both methods got fewer diverged matches as window size increased since larger windows reduce the matching uncertainties. More specifically, geometric constraints are powerful supplements to photo-consistency constraints and help our proposed method, gaining hundreds of converged matches more than the traditional LSM with different window sizes.

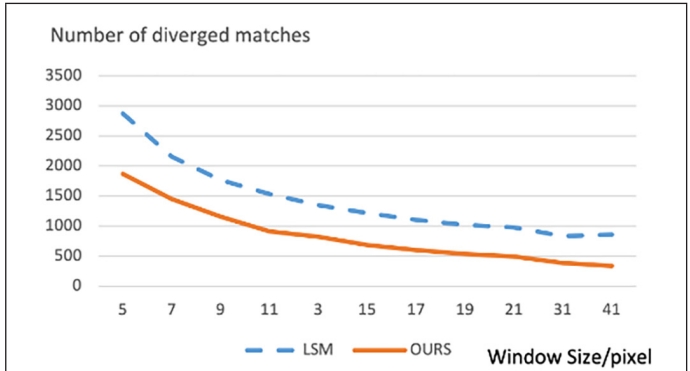

Figure 3. Convergence comparison between traditional least-squares matching and our proposed method. The polylines are the number of diverged matches varying with window size.

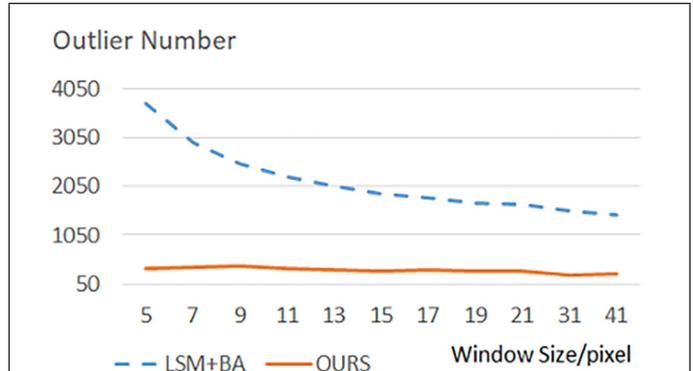

Figure 4. Outlier comparison between the least-squares matching technique before bundle adjustment and our proposed method. The polylines are the number of outliers varying with window size.

### Outlier Comparisons

A series of matching windows were utilized in the LSM + BA and the proposed method for matches corrections, and then outliers in the corrected matches were detected via the following steps: (1) computing 3D coordinates of matches via forward intersection, (2) computing their reprojected image coordinates on all available images via backward projection, (3) computing residual errors on all available images, and (4) counting a view observation as one outlier when the residual error on this view was larger than a predefined threshold (two pixels in our experiments). The outlier numbers of LSM + BA and the proposed method are shown in Figure 4.

Figure 4 shows a fairly intuitive conclusion that the number of outliers is negatively correlated with the size of the windows, as smaller windows tend to feed fewer observations to the LSM. In contrast, our method, due to the employment of the geometric 2D and 3D consistencies, is more robust and shows significantly fewer outliers with all window sizes. This has provided competing advantages over the traditional methods since window size is often a hyperparameter and our method, being less dependent on window size, will yield better robustness and thus can be favorably considered in operations.

### Analysis of Internal and External Accuracy

To comprehensively compare BA (with SIFT), LSM + BA, and our proposed method, we analyzed the internal accuracies of matches/corrected matches and the external accuracies of manually selected points. The hyperparameters of the SIFT algorithm remain the same throughout the experiment (the threshold of the ratio of the distance between the best matching keypoint and the distance to the second-best one is set to 0.35, while other parameters are same as in Lowe [2004]). We selected a total of 33 points with the six-ray connectivity for the accuracy checking, as shown the yellow circles in the satellite image on April 2 in Figure 1. The accuracies of BA, LSM + BA, and our proposed method on the SIFT match set are shown in Figure 5.

We have collected 4210 SIFT matches for testing. It was noted that SIFT, with its nature of locating the interest point in the scale space, yields lower positioning accuracy than corner-based operators (Remondino 2006). Thus, the pure BA result on the SIFT match set, as shown in Figure 4, had lower accuracy than the other two methods in terms of both internal and external accuracies, and its RMSE was larger than one pixel.

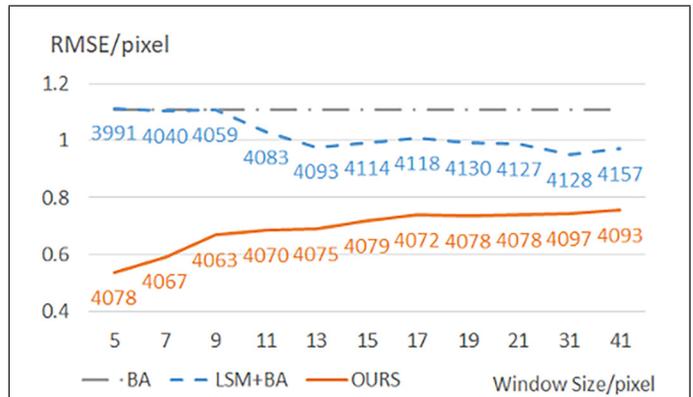

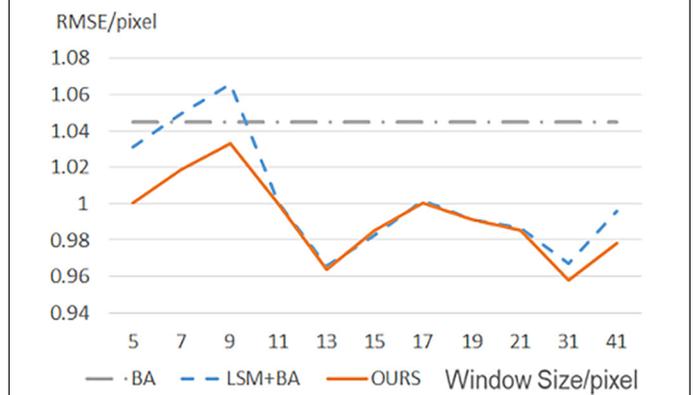

Figure 5. Accuracy comparisons of the three methods on the scale-invariant feature transform match set. The x-axis represents the different matching window sizes, and the y-axis represents the average root-mean-square error of reprojection errors. The numbers around the polylines are the corresponding number of corrected matches.

However, by introducing the LSM to improve the position accuracy for matches, both the LSM + BA method and the proposed method were able to achieve higher internal accuracy, especially subpixel accuracy when the window size was larger than 11×11 pixels. In addition to the photo-consistency constraints, our proposed method also introduces geometric constraints to guide the matching; thus, more matching uncertainties were further reduced compared to the LSM + BA method, and the internal accuracies of our proposed method were always higher than those of the LSM + BA method regardless of matching window sizes. More specifically, our method outperformed the other two methods in terms of internal accuracy regardless of window size.

Figure 5a shows the fittings between the matches/corrected matches and the orientation results, while the orientation results with the highest internal accuracies may not be the best fit for the remaining pixels. Therefore, we manually selected several matches to evaluate the orientation results of different methods, as shown in Figure 5b. The external accuracies of the LSM + BA method sometimes was lower than the pure BA method when the matching window size was small due to high matching uncertainties in the small window size making the LSM unstable. With the increasing matching window sizes, the external accuracies of LSM + BA method surpassed BA by involving more image information. The LSM + BA performed quite closely to our proposed method and was able to achieve subpixel external accuracy when the matching window size was larger than 11×11 pixels. However, our proposed method, integrated with geometric constraint multi-view LSM, was able to achieve the most accurate BA results regardless of window size.

## Conclusions

In this article, we propose a unified framework of feature matching and BA for more accurate orientation results. In general, we formulate the union as the optimization of a global energy function and propose a comprised solution by breaking the optimization into two-step sub-optimizations in an incremental manner. Tested on six full-scale high-resolution satellite images, our comparative study demonstrated that compared to SIFT + BA and LSM + BA, our proposed method is able to consistently achieve the best internal and external accuracies on multi-view satellite image data sets, is much less dependent on window size, and is able to improve the convergence in LSM. Although with a relatively marginal improvement, by simultaneously adjusting both the matches and the biases iteratively, our proposed method plays a role in potentially improving standard "feature + BA" solution.